# Deep Learning Meets OBIA: Tasks, Challenges, Strategies, and Perspectives

Lei Ma, Ziyun Yan, Mengmeng Li, Tao Liu, Liqin Tan, Xuan Wang, Weiqiang He, Ruikun Wang, Guangjun He, Heng Lu, Thomas Blaschke

***Abstract*—Deep learning has gained significant attention in remote sensing, especially in pixel- or patch-level applications. Despite initial attempts to integrate deep learning into object-based image analysis (OBIA), its full potential remains largely unexplored. In this article, as OBIA usage becomes more widespread, we conducted a comprehensive review and expansion of its task subdomains, with or without the integration of deep learning. Furthermore, we have identified and summarized five prevailing strategies to address the challenge of deep learning's limitations in directly processing unstructured object data within OBIA, and this review also recommends some important future research directions. Our goal with these endeavors is to inspire more exploration in this fascinating yet overlooked area and facilitate the integration of deep learning into OBIA processing workflows.**

## I. Introduction and Motivation

DEEP learning has achieved tremendous success in the field of remote sensing over the past decade, particularly in pixel-based and scene-based remote sensing image analysis [1], [2], [3]. However, different processing units generally categorize remote sensing image analysis into three processing paradigms: pixel-based, object-based, and scene-based [4]. Among these, Object-Based Image Analysis (OBIA) is a paradigm that combines pixels with similar semantic information to form segmented objects (vector patches or polygons) [5]. It traditionally uses spatial analysis and machine learning algorithms for the following identification and analysis of these objects [6]. The tasks following the segmented objects mainly include feature extraction [7], classification [8], change detection [9], time series analysis [10], and so on.

However, the conventional OBIA paradigm of "segmentation-feature extraction-classification" has faced a challenge in its usage in deep learning techniques. Deep learning methods like Convolutional Neural Networks (CNNs) struggle to directly handle the unstructured segmentation objects that possess irregular characteristics, not like regular pixels or patches [11]. Therefore, the irregular nature of OBIA's analysis units necessitates the development of various new strategies to apply deep learning methods that require regular patch inputs in OBIA [12], [13]. Deep learning, on the other hand, is even beginning to challenge the traditional OBIA paradigm. For example, DL methods can work without the usual OBIA process and let you get labeled and clustered classification results directly, most of the time in the field of semantic segmentation [14]. This prompts us to reconsider what the general tasks in OBIA are, as well as how it achieves consistent results under advanced techniques.

Here we also searched the literature regarding deep learning and OBIA, and it seemed that OBIA is missing the feast of deep learning prosperity, due to the fact that deep learning is applied relatively late and less and less in the tasks related to OBIA (Fig. 1), and still faces many challenges. To the best knowledge of the authors, there is no comprehensive review summarizing the integration of DL and OBIA, which severely hampers the application of DL in the OBIA domain, due to existing problems. Therefore, it is necessary to write a review article that summarizes the opportunities and challenges of current deep learning in Object-Based Image Analysis, in order to promote the development of DL in the OBIA field.

This article aims to stimulate further research in this valuable yet underutilized research field. Therefore, in Section II, we first summarize and expand upon the general tasks in OBIA following the work of Chen et al. [15], and provide specific definitions for these ununified sub-fields in OBIA. For Sections IV to VII, we outline how current deep learning methods accomplish these tasks within the OBIA framework in order to well guide the research community in adopting deep learning for OBIA. Here, we not only outline the challenges faced by

This work was supported in part by the National Natural Science Foundation of China under Grant 42171304, in part by the Key Laboratory of Land Satellite Remote Sensing Application, Ministry of Natural Resources of the People's Republic of China under Grant KLSMNR-K202301, in part by the Open Research Fund of the State Key Laboratory of Space–Earth Integrated Information Technology under Grant SKL_SGIIT_2024030

L. Ma, Z. Yan, L. Tan, X. Wang, W. He are with School of Geography and Ocean Science, Nanjing University, Nanjing 210023, China (e-mail: maleinju@gmail.com).

M. Li is with Key Lab of Spatial Data Mining & Information Sharing of Ministry of Education, Academy of Digital China (Fujian), Fuzhou University, 350108 Fuzhou, China.

T. Liu is with College of Forest Resources and Environmental Science, Michigan Technological University, Houghton, MI 49931, USA.

R. Wang, G. He are with Beijing Institute of Satellite Information Engineering, Beijing 100095, China.

H. Lu is with State Key Laboratory of Hydraulics and Mountain River Engineering, Sichuan University, Chengdu 610065, China.

T. Blaschke is with Department of Geoinformatics, University of Salzburg, 5020 Salzburg, Austria.

deep learning (DL) in various tasks within OBIA but also provide an overview of existing strategies to address these challenges.

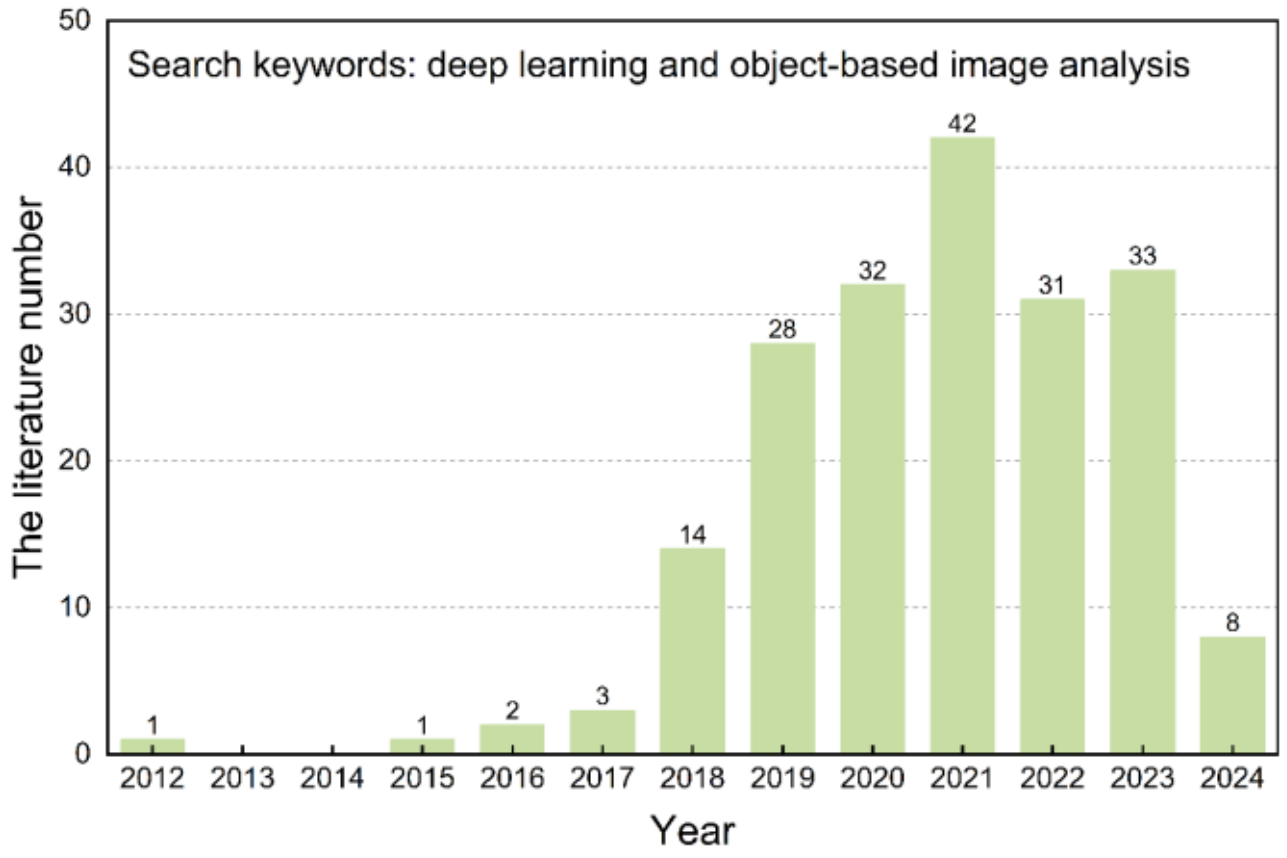

Fig. 1. The trend of publications related to DL and OBIA (searching time: 2024.4.8).

## II. General tasks in OBIA

In this section, we briefly summarize the core tasks relevant to Object-Based Image Analysis, which have traditionally included segmentation, classification, and feature extraction [15] and recently have broadened OBIA's scope to include time series analysis and change detection, reflecting advancements in state-of-the-art research for object-level remote sensing. Furthermore, the parameter inversion task has become prominent in OBIA because of its applicability in estimating various earth surface parameters, and this paper also redefines the object detection mission to focus on identifying specific geographic objects of interest within the imagery. Subsequently, we delineate seven specific tasks that are currently central to OBIA:

### *A. Segmentation*

Segmentation is a crucial part that distinguishes the object-based paradigm from the pixel-based paradigm. Traditional unsupervised segmentation algorithms used for OBIA mainly fall into two categories: 1) edge-based segmentation algorithms and 2) region-based segmentation algorithms [16]. Semantic segmentation algorithms based on deep learning models have also become a mainstream research direction in segmentation in recent years, except for unsupervised segmentation, since they allow for direct labeling of the segmentation results without the need for additional feature extraction and classification processes [17].

Edge-based algorithms in image segmentation identify and close edges to delineate objects, starting by detecting sudden changes in pixel properties as indicators of edges [18], [19]. Edge detection can be divided into three steps: filtering, enhancement, and detection [20]. Each of these steps has been the focus of extensive research, leading to a variety of innovative approaches, including methods like Sobel [21], Canny [22], and Watershed Transformation [23], all adept a popular strategy at identifying edges through local gradient operators. However, edge detection often results in incomplete or broken edges which makes it challenging to transform identified edges into closed boundaries [24], [25]. Although many advanced methods like the Hough transform [26] and neighborhood searches [27] are employed to exclude noise-induced edges, connect gaps, and unify edge segments corresponding to single objects, identifying perfect edges for delineating objects remains difficult.

Region-based algorithms take an opposite approach compared to edge-based algorithms, as they search an initial region and then detect edges. These methods can be divided into region-growing methods (bottom-up strategy) and region-splitting (up-bottom strategy) methods [16]. Region-growing methods often rely on seed selection and similarity criteria to determine the growth and merging of regions [25]. After seed selection and localization, the region grows by merging adjacent pixels to obtain an object, according to a specific homogeneity criterion [28]. Except for growing based on seeds, some algorithms, like Multi-Resolution Segmentation (MRS) [29], Mean-Shift (MS) Segmentation [30], and Fractal Net Evolution Approach (FNEA) [31], merging objects to a bigger object with local criteria and global criteria, can also be called region-merging techniques as the most widely used method for OBIA. Region-splitting methods split the image into segments with inhomogeneity criteria such as grey values or texture [32]. These methods iteratively split the image into smaller subregions, until all subregions are homogeneous. Sometimes, we can use the region merging approach to merge similar subregions, as split subregions may be too square or fragmented [33]. However, for all region-based algorithms, challenges remain in defining appropriate parameters and adapting to different image settings. Compared to edge-based algorithms, they are insensitive to object boundaries. Thus, some scholars have utilized a hybrid strategy to overcome the limitations of both edge- and region-based methods [34]. Though these methods obtained better results, the processes are troublesome.

The semantic segmentation algorithm differs from the aforementioned algorithms because semantic segmentation categorizes each pixel in an image into a predefined class, which allows it to complete the OBIA workflow independently (e.g., classification). Early semantic segmentation algorithms were mostly based on classic machine learning models, such as Markov Random Fields, which model the relationships and interactions between neighboring pixels to enhance prediction accuracy [35]. However, these methods relied heavily on statistical assumptions and handcrafted features, which somewhat limited their effectiveness and adaptability to complex scenarios. In recent years, with the advancement of deep learning technology, semantic segmentation models have made significant progress, even giving rise to instance segmentation and panoptic segmentation [36], [37]. For example, there are big needs for the mission of counting trees by instance segmentation [38], [39]. Therefore, segmentation including sematic and instance should also be the important exploration direction in the field of OBIA while many studies treat them as independent tasks.

### B. Classification

Object-Based Image Classification (OBIC) has become a pivotal focus in the field of OBIA due to its ability to effectively incorporate spectral, geometric, textural, and other characteristics of ground features during the classification stage. In the past few years, advances in remote sensing technology have also sped up the evolution of OBIC. It has gone from using unsupervised rule-based methods [40], [41], [42] to using supervised methods like machine learning (ML) algorithms [43], [44], [45] and deep learning (DL) techniques [11], [46], [47].

In its early stages, OBIC relied on expert knowledge and experience to set thresholds using rule-based classification methods. These methods were effective for data with clear boundaries and deterministic features. However, their utility was limited when dealing with the uncertainty and fuzziness common in real-world data. To address this, researchers like Mathieu et al. [41] and Jacquin et al. [48] introduced fuzzy rule-based classification methods that could handle the uncertainty inherent in segmentation categories, and the developed rule sets demonstrated significant portability [42], [49]. Despite this, the ambiguity in class assignment by fuzzy classification necessitated the use of defuzzification techniques to achieve clearer and more reliable classifications [50].

With ongoing advancements in remote sensing technology, fuzzy rule-based classification began facing limitations, prompting a shift towards supervised classification methods like Decision Trees (DT), Support Vector Machines (SVM), and Random Forests (RF). These methods have seen rapid development and widespread application within OBIC, as evidenced by studies from Duro et al. [51], Wu et al. [52], and Qiu et al. [53]. Many researchers, including Li et al. [8] and Wang et al. [44], have employed multiple machine learning (ML) methods, comparing their performance. Peña et al. [43] notably assessed the effectiveness of DT, SVM, Logistic Regression (LR), and Multilayer Perceptrons (MLP) in both standalone and hierarchical classifier systems, finding that hierarchical approaches significantly enhanced the accuracy for the least accurately classified classes. Additionally, studies like Radoux et al. [54] have concentrated on refining sampling strategies to improve the precision of Land Use and Land Cover (LULC) classification. Furthermore, the importance of feature selection has also been recognized as a critical step in enhancing the accuracy of object-based classification [51], [55]. Yet, despite these advancements, consensus on the overall effectiveness of integrating feature selection methods with OBIC remains elusive, though Ma et al. [56] have made notable contributions to bridging this gap.

While ML algorithms have recorded considerable success in OBIC, their performance largely hinges on factors such as sampling strategies, feature extraction and selection, and classifier parameter tuning, as highlighted by Radoux et al. [54], Ma et al. [56], and Li et al. [8]. More recently, deep learning (DL) has shown great promise in OBIC applications due to its end-to-end processing and robust feature extraction capabilities, as explored by Ma et al. [57]. However, the requirement for fixed-size inputs in DL poses a challenge due to the irregular nature of actual objects, making it difficult to integrate DL with OBIC [11], [58]. Despite initial studies attempting to address these challenges, a unified approach and consensus within the academic community remain to be established. Consequently, Section IV of this paper is summarized so as to contribute to this field.

### C. Change detection

Object-based change detection employs image objects as the primary units of analysis, significantly enhancing accuracy by taking into account the objects' shape, texture features, and spatial context. Depending on the strategies for object acquisition, object-based change detection methods can be divided into three categories [59]: 1) image-object overlay change detection; 2) multi-temporal object change detection; and 3) object-to-object change detection.

Both the image-object overlay and multi-temporal object strategies aim to produce segmented objects that are topologically consistent. The image-object overlay approach performs segmentation on a single image, then applies the segmented results to additional images [60], [61]. Despite its widespread acceptance due to its simplicity and computational efficiency, this method falls short in capturing the dynamic boundaries of objects due to under-segmentation leading by limited information for segmentation. On the other hand, the multi-temporal object strategy involves stacking all images together for segmentation, thereby fully utilizing the information from all temporal phases [62], [63], [64]. Its distinct advantage lies in effectively obtaining static and dynamic boundaries, but it may encounter the issue of over-segmentation. Subsequently, feature vectors—spectral, textural, or ecological indices from remote sensing—are extracted for each object at different times. These vectors are compared to quantify the magnitude of changes, which is then classified through thresholding or clustering to produce binary change detection maps [9], [60], [65], [66]. For both strategies, due to the time-consistent nature of objects, it is difficult to identify changes in objects through changes in their geometric characteristics.

Object-to-object comparison entails segmenting each image independently, which precisely delineates the boundaries of objects across different time periods and facilitates the use of geometric information for change detection. However, without geometric position constraints, establishing spatial correspondences between objects is crucial and challenging. Typically, the association confidence between objects is determined based on the spatial distance between centroids, the degree of overlap, and the alignment of object coordinates [67]. Once object linking is successfully achieved, it is possible to recognize the change by revealing changes in geometric properties such as area, shape, and quantity [68], [69]. Furthermore, by comparing further classification results, a "from-to" class label change matrix can be generated [70], [71].

### D. Parameter inversion

One fundamental characteristic of remote sensing in the 21st century is the extensive use of quantitative algorithms for inverting Earth surface parameters [72]. Classically, parameter inversion task aims at estimating quantitative parameters from radiance data in remote sensing images across various fields such as biology [73], [74], hydrology [75], surface environment [76], urban landscape [77], and so on. Object-Based Image Analysis (OBIA) distinguishes itself from conventional pixel-based methods by utilizing image objects from segmented images, rather than individual pixels, as the primary unit of analysis. This shift offers enhanced flexibility and adaptability across different scales, regions, and specific parameter inversion tasks [15].

OBIA has proven particularly effective in fields requiring precise parameter inversion. In biology, for example, the estimation of aboveground biomass (AGB) has gained significant attention due to its relevance to ecosystem and global change studies. Studies by Kajisa et al. [78] and Silveira et al. [79] using Landsat-based OBIA for AGB inversion have shown superior outcomes compared to pixel-based approaches, notably improving spectral homogeneity and reducing positional discrepancies between image and field data. The focus on AGB inversion has recently shifted towards data fusion, incorporating both optical and synthetic aperture radar (SAR) data using an object-based approach with multi-sensor images [80], [81]. Additionally, the integration of radiative transfer models (RTM) with OBIA has proven effective in inverting plant parameters such as normalized difference vegetation index (NDVI), leaf area index (LAI), leaf chlorophyll (LC), plant density (PD), green vegetation cover fraction (FCV), and crop evapotranspiration (ETc) [82], [83], [84], [85], [86].

OBIA is also widely applied in inverting parameters for soil, geomorphology, and glaciers. Since 2018, Attarzadeh et al. [87] have developed an OBIA methodology for soil moisture content (SMC) inversion by coupling SAR and optical data, enabling multi-scale SMC analysis using the Random Forest (RF) algorithm and support vector regression (SVR) with multi-sensor images. In geomorphology, OBIA has been employed to map parameters and associated changes, such as digital elevation models (DEMs), particularly in geohazard assessments [88]. For glaciers, Raghubanshi et al. [89] used an OBIA approach for the normalized difference snow index (NDSI), effectively distinguishing snow from water.

Recently, urban environments have increasingly benefited from OBIA for parameter inversion. The method effectively models spatial relationships between objects, considering the spatial structure and morphological features of the urban landscape through indices like total edge (TE), edge density (ED), mean patch edge (MPE), mean patch size (MPS), and mean shape index (MSI) [77], [90], [91]. Looking forward, the adoption of advanced deep learning models in OBIA for diverse fields is anticipated to further enhance parameter inversion capabilities.

### E. Time series analysis

Satellite Image Time Series (SITS) has become an invaluable resource for studying dynamic changes on Earth's surface, offering an enhanced capability to track terrestrial features over time [92] to support the tasks of classifications, change detection, and even segmentation. Unlike single-observation imagery, SITS incorporates multi-temporal data, providing a more comprehensive view of environmental dynamics. While much of the research in SITS analysis has traditionally focused on pixel-level analysis [10], the growing clarity offered by advances in sensor resolution has highlighted the benefits of object-based analysis [93], [94]. Recent studies have started to delve into the approach of object-based SITS, classifying existing object-based SITS analysis methods into three categories: 1) Similarity measurement methods; 2) Model-based methods; and 3) Machine Learning (ML) methods.

Similarity measurement methods usually use raw time series data to construct the time series of certain features or indexes (e.g., NDVI), effectively capturing the phenological variation patterns of different land types to support their classification. An example is the object-based Time-Weighted Dynamic Time Warping (TWDTW) used by Belgiu and Csillik [10] for agricultural field mapping, which demonstrated superior performance over its pixel-based counterpart. Csillik et al. [95] further refined this approach with the object-based Time-Constrained DTW, achieving more accurate classification by eliminating incorrect matching paths. Most research in this area has focused on DTW, but He et al. [96] suggested further exploration of other approaches, such as shape-based distance (SBD) and global alignment kernel (GAK), and it was found that GAK measure usually works better for agricultural mapping than the others.

Predicated on statistical assumptions, model-based methods generally aim to achieve change detection by modeling dense time series, effectively mitigating issues like noise and lighting variations [97]. Successful pixel-based models include LandTrendr, BFAST, and CCDC [98], [99], which, despite their successes, often miss subtle disturbances influenced by factors like topography, climate, or human activities. To address this, some researchers have turned to segmented object analysis, utilizing the spectral, morphological, and textural characteristics of objects to enhance SITS analysis outcomes. This approach faces the challenge of maintaining consistent object segmentation across spatial and temporal scales due to the inherently dynamic nature of landscape boundaries in SITS [93], [100]. Recent studies have made progress by extracting objects post-pixel-level change detection, somewhat alleviating the challenges associated with using segmented objects in dense SITS analysis [93].

However, model-based methods are not without their own challenges, including complexity in modeling and a heavy reliance on model accuracy. In recent years, the integration of Object-Based Image Analysis (OBIA) with ML algorithms has gained traction due to their simplicity and robustness in automatically extracting complex features from data. Notable applications include Random Forests [101], Support Vector Machines [102], and k-Nearest Neighbors classifiers [103]. For example, Zhang et al. [104] used an object-based Random Tree method with MODIS data to extract large-area wetland information, while Cai et al. [101] combined Sentinel-1 and MODIS images to optimize feature extraction for paddy rice using an object-based RF classifier. Adeli et al. [105] successfully employed SVM and RF classifiers with NISAR dense time series for wetland classification, showcasing high accuracy.

Looking forward, the potential of Deep Learning (DL) in handling time series data appears promising due to its ability to automate learning processes without extensive expert input and its computational efficiency [106]. Despite its widespread application in other areas, such as Recurrent Neural Networks (RNNs) for sequence analysis, the integration of DL with object-oriented SITS analysis is still in its infancy. There is a significant opportunity to further explore DL applications within OBIA for SITS analysis [57], potentially opening new avenues for understanding and managing Earth's dynamic landscapes.

### *F. Object detection*

With the increasing resolution of satellite images, OBIA has emerged as a novel paradigm for object detection in remote sensing monitoring. Unlike the detection of a single target from a photo in the computer field, object detection here is defined as the extraction of individual land cover type information using the OBIA technique.

That is, OBIA focuses on detecting specific categories of geographical objects (e.g., landslides, glaciers, and buildings), treating objects as evaluation units rather than pixels to assess detection accuracy. Typically, OBIA-based object detection methods involve two steps: image segmentation and object-based classification. By surpassing the limitations of pixel-based methods, OBIA has gained widespread application in geoobject detection [15]. According to survey findings, research on object detection using OBIA predominantly centers on landslides and glaciers [107], [108], [109]. Additionally, studies also cover buildings [110], gully erosion [111], and avalanches [112].

In the field of landslide detection, OBIA demonstrates significant potential by integrating spectral, terrain, and texture features into image segmentation and classification processes. Dabiri et al. [113] utilized Landsat images, surface parameters, water temperature parameters, and spectral indices to identify five types of landslide areas. Pawluszek et al. [114] compared OBIA-based methods' effectiveness in detecting landslides in forested and cultivated areas, noting its competence in forested regions but limited efficacy in agricultural areas. Tehrani et al. [115] proposed a global mountain landslide detection methodology using OBIA techniques, despite the geographical constraints of previous OBIA-based studies. Dias et al. [109] discussed challenges in identifying small scars within landslide areas and difficulties in detecting landslide tails. Lu et al. [107] used the built-in landslide sample library, which is based on deep learning and transfer learning in the traditional OBIA paradigm, to get information about landslides. Future research directions may involve further refinement of rule sets and leveraging prior knowledge.

In terms of glacier detection, the primary challenge lies in distinguishing glaciers from surrounding terrain, especially broken ice. Early detection of glaciers concealed by broken ice often relies on manual interpretation. Robson et al. [116] used Landsat 8 optical data, Shuttle Radar Topography Mission (SRTM) elevation data, and synthetic aperture radar (SAR) images to find glaciers near Mount Manaslo, Nepal. They did this by leveraging OBIA's ability to handle data from multiple sources. In 2020, they further employed a combination of deep learning and OBIA for rock glacier detection [108]. This study utilized a convolutional neural network (CNN) to identify recurring textures and generate heatmaps, followed by OBIA for image segmentation and object classification, significantly reducing the workload of inventory creation. Similarly, detecting water bodies faces challenges due to confounding features caused by freezing water into ice during the winter months, necessitating careful consideration of similar landforms in their vicinity. Building upon this premise, Korzeniowska et al. [117] assessed glaciers on the Qinghai-Tibet Plateau using Landsat images and SRTM data.

Through a comprehensive review of existing studies, it is evident that OBIA-based investigations commonly leverage multiple sources of earth observation data, with more information enhancing the accuracy of OBIA methods. Furthermore, the OBIA approach primarily relies on discerning spectral, spatial, and textural disparities between targets and their surrounding features, rendering it particularly efficacious for large-scale target detection. Nevertheless, challenges persist when applying OBIA in scenarios characterized by low resolution or small targets.

### *G. Feature extraction*

OBIA offers meaningful units to describe geographic objects or regions in images, and feature extraction, which typically follows image segmentation, is crucial for processing and interpreting high-resolution imagery. Classic object-based features encompass spectral, geometric (shape), texture, and contextual information, often extracted with specific formulas [15]. More recently, deep learning models have been employed for feature extraction in OBIA, treating a trained neural network as a feature extractor [118].

Spectral, geometric, and texture features are commonly utilized for characterizing specific attributes of individual objects [119], [120]. These features, derived from collections of homogeneous pixels within objects, can be amplified for further tasks like classification. For example, by extracting NDVI values of different objects, vegetation can be distinctly identified. Spectral features often include statistical values of individual bands or the results of mathematical operations between multiple bands [51]. Geometric features may encompass area, aspect ratio, compactness, and smoothness of individual objects [121], [122]. Texture features describe the regular patterns within an object, commonly derived from a Gray-Level Co-occurrence Matrix (GLCM) and include features such as entropy, contrast, correlation, and variance [123].

However, these features primarily focus on the intrinsic characteristics of individual objects and may malfunction when segmentation results are suboptimal. In cases where interpreting objects' connections is necessary, scholars have explored extracting contextual information to compensate for insufficient interaction between objects [15], [124]. For example, Wang et al. [125] hypothesized that within a city, semantic features exhibit regular, block-like patterns at the community level, necessitating the consideration of relationships between objects.

Recent studies have demonstrated that deep neural networks trained on large datasets possess excellent feature representation capabilities [126], [127]. In shallow layers, shallow features like shape and texture can be extracted [126], while deeper layers output more abstract semantic features such as the probability of a particular class [128]. Subsequently, utilizing neural networks as feature extractors can optimize OBIA workflows, especially for providing additional semantic information. This approach has already been implemented to generate new feature data for classification (Section IV B).

## III. Recent advances and challenges in deep learning applied to OBIA

### A. Popular DL models in OBIA

The application of deep learning methods to various tasks within OBIA has gained traction due to DL's network structure, which is well-suited for OBIA tasks. This structure enables the capture of feature information and eliminates the need for extensive feature engineering, thus enhancing OBIA accuracy and efficiency [129]. Among DL models, convolutional neural networks (CNNs) are particularly popular and have been widely applied to OBIA tasks.

Classification and segmentation are two prevalent tasks in OBIA where DL models are extensively utilized (Fig. 2). For classification tasks, CNNs are commonly employed for basic classification, variants such as region convolutional neural networks (R-CNN), deep convolutional neural networks (DCNNs), and fully convolutional networks (FCNs) refine and enhance classification results [13], [130], [131], [132]. In segmentation research, CNNs are often used as a foundation, followed by methods like fully convolutional networks (FCNs) to enhance segmentation accuracy [133], [134], [135]. Extensions such as Mask R-CNN aid fine segmentation by generating accurate pixel-level masks of objects [136], [137].

Object detection and change detection are also gaining attention in DL-based OBIA tasks (Fig. 2). When CNNs detect objects, they effectively handle objects of varying sizes through the use of multi-level convolution [138]. For more complex objects, they use a region proposal network (RPN) to automatically find object boundaries [139], [140]. Regarding change detection, CNNs and fully convolutional networks (FCNs) are capable of meeting the demands of large-scale change detection tasks [141], [142], [143]. These models effectively capture temporal information in multi-temporal images, facilitating the detection of changes over time. Additionally, architectures based on Long Short-Term Memory (LSTM), a variant of Recurrent Neural Networks (RNN), are explored for long-time series research to identify long-term dependencies in time-series images [144], [145], making them suitable for tasks requiring analysis of temporal patterns and trends.

Despite DL's potential for OBIA tasks, the unstructured and irregular characteristics of objects pose challenges for DL models that require regular inputs (e.g., patches) [146]. In addition to CNNs and RNNs, graph-based deep learning models, such as graph convolutional networks (GCNs), are emerging as potential models for various OBIA tasks [147], [148]. GCNs leverage the connections between nodes in the graph structure to model spatial relationships between objects and adapt to irregular object shapes [149], [150]. Furthermore, effective strategies to address the irregularities and inconsistencies of objects for DL in OBIA are ongoing research topics (Section IV).

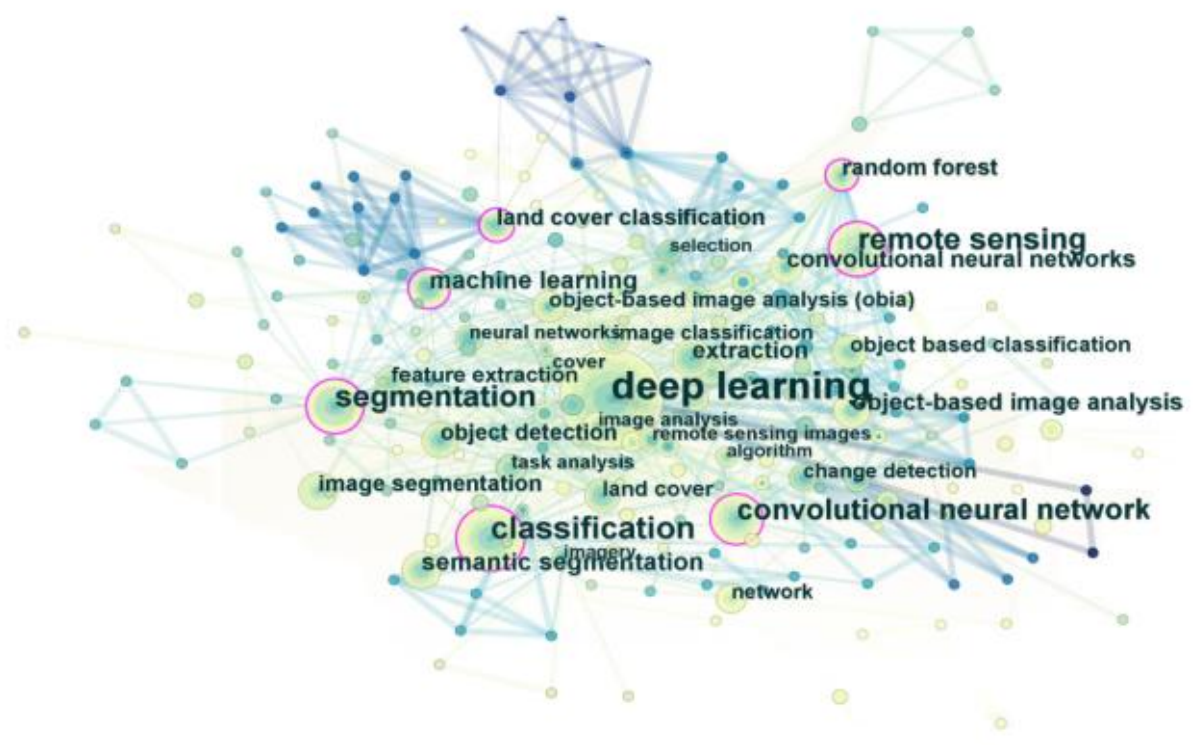

Fig. 2. A visual representation of the frequency of keywords appearing in the peer-reviewed literature searched by the keywords "object-based image analysis and deep learning" (searching time: 2024.4.8).

### B. Possible pitfalls

It's critical to unify the concept and understanding of OBIA before discussing it with deep learning. Currently, the field is muddled with confusing concepts, leading to misguided studies according to some researchers. For instance, Guirado et al. [151] inappropriately compared OBIA with DL methods by using traditional classifiers such as Random Forest (RF) within the OBIA framework and comparing these to standalone DL algorithms. Ideally, OBIA should align with pixel-based classifications, whereas RF should align with DL classification methods. Therefore, here we summarize a few confusing pitfalls in OBIA and DL.

*OBIA as a Paradigm, Not Merely an Algorithm:*

Many studies [151], [152] erroneously treat OBIA as an algorithm, at least conceptually, and compare it directly with DL or other machine learning algorithms. This paper aims to correct this misconception, emphasizing that OBIA is a comprehensive paradigm for analyzing remote sensing data. For example, semantic segmentation using deep learning is considered a subfield of OBIA because it aggregates and labels pixels, similar to traditional OBIA outputs. However, it does not mean that the deep learning models as classifiers could be on par with OBIA by comparing with conventional object-based image classification workflows due to the different output data structures [151], [152]. That is, CNN-based method in the study of Guirado et al. [151] outputs object detection results like points, while the OBIA outputs land cover results. Conversely, Liu et al. [131] offer a careful comparison, treating DL as a classifier alongside traditional algorithms (RF, SVM) within the OBIA framework.

*Expanding the OBIA Framework, without restricting it to traditional workflows:*

With advancements in AI technology and big data, we advocate for an objective-oriented OBIA paradigm. Any geographic spatial analysis method that aggregates similar pixels to derive analytical results should fall under the OBIA umbrella, regardless of the techniques or data types used. This expansion is essential as object-based research has evolved beyond the confines of traditional OBIA workflows with images. For example, Yan et al. [153] introduced OSM data for local climate zones mapping within the OBIA paradigm, not solely relying on remote sensing data. Ye et al. [93] integrated segmentation as part of the initial pixel-based time series process, while Guttler et al. [154] tackled inconsistencies in segmented objects from different temporal images by constructing graph-based networks. This evolution shows that time series analysis has become a crucial component of OBIA.

*Conceptual Misunderstandings in DL Integration into OBIA:*

Deep learning's applicability across various stages of remote sensing image processing [57] can lead to misunderstandings. When DL is employed as a classifier, it should adhere to the methodologies outlined by Liu et al. [131], which are detailed further in Section IV of this paper. When used as a segmentation tool, its role depends on whether it provides class labels. If segmentation does not yield labels, it can be a step within the traditional OBIA framework; otherwise, semantic segmentation that offers labels might constitute a complete object-based classification process. Therefore, in Section IV C, we include semantic segmentation using DL algorithms as a key OBIA classification strategy with DL integration. We hope such clarifications contribute to the expansion of OBIA's paradigm.

### *C. Key challenges*

Due to the inherent complexities of OBIA and deep learning, applying deep learning to OBIA requires addressing several challenges. Here, we outline four critical issues where conflicts and potential improvements emerge:

*Inconsistencies in Analysis Units:*

Here we define the inconsistencies in two dimensions for the basic units: spatial and temporal. Spatially, the segmentation objects produced by algorithms are irregular, posing challenges for object-level image classification. However, over the past few decades, researchers have made significant theoretical and methodological advancements in this area [56]. For deep learning, the requirement for regular patches as inputs [11] complicates direct integration of DL into the OBIA framework for handling inconsistently segmented objects. Attempts have been made to develop suitable integration strategies, but these often suffer from low efficiency. Temporally, the greatest challenge in time series analysis is the inconsistency of segmented objects across different times, which leads to many studies using consistent objects from one or several images for object-based time series analysis, often overlooking actual changes over time.

*Insufficient Training Samples:*

Traditional OBIA processes segmented polygons, effectively compressing the original remote sensing data, which reduces both data volume and computational requirements. This was suitable for the initial rule-based OBIA, and then, despite the integration of machine learning, the lack of training samples for OBIA did not significantly draw scholars' attention. However, deep learning poses a major challenge due to its demand for large training samples, which is difficult to meet within the OBIA paradigm. Additionally, the uncertainty of segmentation and scale differences among segmented objects make constructing object-based benchmark datasets extremely challenging. Therefore, sample augmentation methods appear necessary for OBIA with DL [45].

*Balancing Boundary Blur and Classification Accuracy in Semantic Segmentation:*

Fully Convolutional Networks (FCN), as a typical semantic segmentation model, extended deep classification networks to segmentation tasks [155], which was different from the old OBIA framework because it labeled pixels that had similar features except for spatial aggregation of these pixels. However, semantic segmentation still faces the mainstream dilemma of balancing classification performance with boundary delineation. This means a series of improvements in practical applications is necessary to ensure precise boundary information while also identifying land cover types well [156], [157], [158]. For instance, the popular FCN in the OBIA domain often loses detail due to its large receptive field, leading to blurred segmentation boundaries [128], which makes additional post-processing essential.

*Poor Interpretability:*

Initially, OBIA favored rule-based methods for classifying segmented objects, which provided strong interpretability [159]. Despite widespread use of machine learning in OBIA and extensive research in feature selection to enhance understanding of the classification process [56], [160], deep learning inherently suffers from poor interpretability issues [161]. This lack of transparency is a significant barrier to DL adoption in the OBIA domain.

## IV. EXISTING STRATEGIES FOR APPLYING DEEP LEARNING TO OBJECT-BASED REMOTE SENSING IMAGE CLASSIFICATION

TABLE I

THE SUMMARY OF PROS AND CONS FOR SEVERAL STRATEGIES OF CLASSIFICATION INTEGRATING DEEP LEARNING WITH OBIA

| Strategy | Subtypes | Description | Pros | Cons | Examples |
|---|---|---|---|---|---|
| **Strategy 1:** Linking the type of the classified patch with the segments | Bounding boxes for patches | Resize bounding boxes and then add padding or mask | All objects are uniformly processed, making the workflow simple | The resize operation will result in the scaling of the original image | Jozdani et al. [162] |
| | Candidate points for patches | Use algorithms to search candidate points and then extract adaptive patches | The size and shape differences between objects have been well-considered | The workflow is complex, requiring multiple branch networks to handle patches of different sizes | Zhang et al. [168]; Martins et al., 2020 [12] |
| **Strategy 2:** Deriving new data by deep learning as inputs for object-based classification | Derived data as feature inputs for classification | Utilize data derived from deep learning models to extract object-level features | Features derived from deep learning can provide semantic or class-specific information | Deep learning models require extensive data for training and this strategy makes OBIA workflow more complex | Thomas et al. [170] |
| | Derived data as original inputs for segmentation | Utilize data derived from deep learning models to assist segmentation | Unlike spectral bands, data derived from deep learning contains less noise | The data derived from deep learning may lack boundary information | Ghorbanzadeh et al. [37]; Timilsina et al. [128] |
| **Strategy 3:** Identifying the land cover directly by segmentation | NA | Using semantic segmentation directly produces homogeneous image regions corresponding to image objects in traditional Object-Based Image Analysis (OBIA), each assigned to specific classes | Semantic segmentation with deep learning simplifies the traditional OBIA process to accomplish segmentation and labeling simultaneously, and it increases classification accuracy | There is already a segmentation boundary blur issue due to the large receptive field of deep learning, which generally requires post-processing to get a fine boundary | Timilsina et al. [128]; Lin et al. [179] |
| **Strategy 4:** Pixel- or patch-based classification by deep learning, and then refine the classification by segments | Initially classification using patch-based classification | Employ deep learning models (e.g., DCNNs) for the classification of input image patches, and then refines the classification by traditional segmentation (e.g., MRS) | Leverage the high classification accuracy of deep learning and the precise boundary delineation of traditional segmentation objects to overcome the boundary blurring issues inherent in semantic segmentation in strategy 3. | The classified patches may not align with the boundaries of the segmented objects, and the integration of classification and segmentation processes proves too complex | Sun et al. [189]; Tong et.al. [185]; Pan et al. [190] |
| | Initially classification using using pixel-based classification | Initially perform pixel-wise classification (e.g., semantic segmentation), and then refines the classification by traditional segmentation (e.g., MRS) | Utilize the pixel-wise classification approach to ensure consistency with segmentation objects, enabling precise categorization. Integrate it with traditional OBIA techniques for accurate boundary delineation to | Pixel-by-pixel classification is an inherently computationally intensive task, with spatial redundancy contributing to the high computational costs | Liu et al. [131]; Du et al. [191] |
| | | | surmount the boundary ambiguity issues inherent in the process of semantic segmentation | | |
| **Strategy 5:** Graph-based deep learning for OBIA | Graph-based method for spatial data | Construct an undirected graph with nodes and edges, where each node represents an object and edges connect adjacent spatial objects. Land cover mapping is usually treated as a node classification | The graph-based approach excels in its exceptional flexibility, enabling it to handle complex objects of varying sizes, and thus higher accuracy is achieved by leveraging the contextual relationships between objects | Need to integrate feature extraction methods, making the method more complicated | Liu et al. [201]; Dufourg et al. [194] |
| | Graph-based method for temporal data | Construct a directed graph with nodes and edges, where each node represents an object and edges connect adjacent time objects. Graph-level analysis is required to deal with the nodes and edges | Handle complex objects of varying times, and enable it to depict dynamic changes in objects over time | Graphs used are usually directed and more complex | Khiali et al. [147]; Debusscher et al. [203] |

### A. Linking the type of the classified patch with the segments

The hardest part of classification tasks that use Convolutional Neural Networks (CNNs) and irregular geo-objects is turning these objects into CNN input patches of a fixed size. Current research predominantly relies on two strategies for patch extraction, as shown in Fig. 3: 1) Identifying the bounding box of an object and converting it into a fixed-size patch; 2) Identifying one or several points representing the object and extracting fixed-size patches centered around these points.

The strategy of extracting patches through bounding boxes involves first identifying bounding boxes and then optimizing the information contained within these boxes (Fig. 3 a). As the bounding box represents the minimal enclosing rectangle of a segment, this strategy accounts for variations in shape and size across segments. The main problem here is finding and adjusting bounding boxes. While many studies directly use the image coordinate system to orient bounding boxes [131], [162], Zhang et al. [90] argued that there could be countless enclosed bounding boxes in arbitrary orientations for a given object. To address this, they proposed extracting the moment bounding box that best represents the segment by leveraging the moment of inertia. Additionally, some scholars have attempted to optimize the size of bounding boxes. Majd et al. [163], for instance, eschewed the extraction of enclosed bounding boxes and instead analyzed the homogeneity of the surrounding information of the segment. If the surrounding information was homogeneous, they enlarged the size of the bounding box to provide more contextual information.

Once bounding boxes are defined, further conversion into fixed-size patches becomes necessary due to size variations among bounding boxes [164]. Existing studies typically adopt specific strategies to optimize the information within bounding boxes. The most common approach involves using the bounding box to crop the image and then resizing it to the required size [131]. However, directly resizing the cropped image may alter the aspect ratio, so some studies have opted to employ zero padding to fill the patches [162], [165]. On this basis, further choices are made regarding whether to mask non-object information [164].

The strategy of extracting patches through candidate points (Fig. 3 b), also known as convolutional positions [12], [166], involves first identifying optimal points and then determining the size of the patches. Consequently, these studies often concentrate on selecting better candidate points and, when necessary, comparing the classification effects of patches of different sizes to determine the optimal size. The most common practice for segment classification is to choose the center of an object and extract one patch representing one segment [11], [167]. However, this approach overlooks the shape and size variations of segments. Hence, some studies have attempted to search for several candidate points and extract patches of different sizes. Various approaches can be employed for point searching. Zhang et al. [168] utilized a moment bounding box to find candidate points based on its axes. Martins et al. [12] utilized the skeletonize algorithm, which generates the morphological representation of segments, to search for optimal points. Lv et al. [166] cropped segments into sub-segments using the binary tree sampling method, enabling the identification of candidate points for each sub-segment.

Additionally, determining the optimal patch size becomes challenging due to variations in segments. Several studies have tackled this challenge by proposing different approaches. Zhang et al. [168], for instance, categorized objects into two types: general objects and linear objects. They employed a large input-based CNN for general objects and a smaller input-based CNN for linear objects. Similarly, Martins et al. [12] addressed this issue by extracting multi-scale patches, analyzing objects for different targets, and subsequently designing a multi-branch CNN, all of which contributed to enhancing the final prediction.

Current methods employ various strategies, but there hasn't been a comprehensive comparison of the effects of different patch extraction methods on object-based classification. We contend that the choice of method should align with the task objectives. For example, some researchers advocate for patches closely matching the attributes of the object to avoid introducing heterogeneous information [168], while others argue that a simple bounding box is too small and lacks sufficient information [163], suggesting an expansion of the bounding box. Given these differing perspectives, there is a need for a comprehensive comparison of these patch extraction methods.

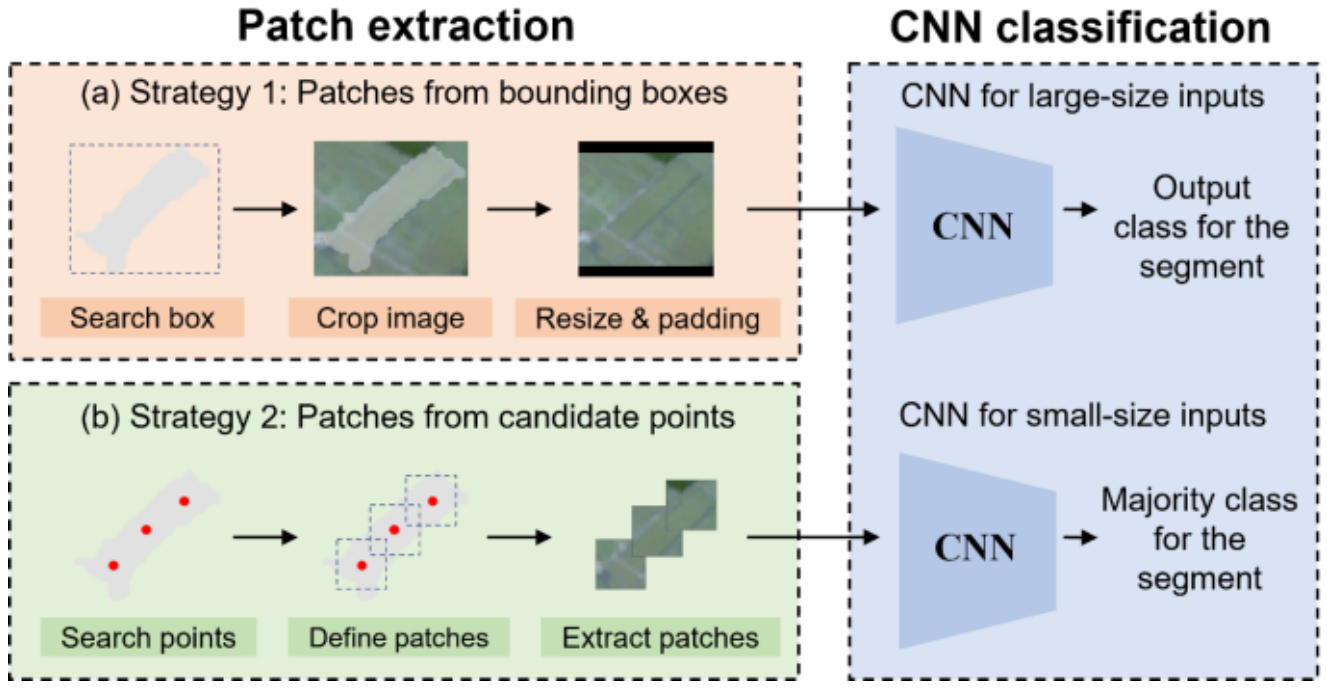

Fig. 3. Clasifying segments by patch-based CNN.

### B. Deriving new data by deep learning as inputs for object-based classification

Deep learning models have demonstrated remarkable capabilities in feature representation [126], particularly in capturing high-level information that can generate new data for the object-based classification workflow. Consequently, this strategy typically aligns with the traditional object-based classification framework and utilizes derived data to support the classification or segmentation process, as illustrated in Fig. 4 [127], [128].

Deep learning-derived gridded data can be effectively utilized for object-based feature calculation, directly contributing to the classification process. Traditional methods typically extract meaningful object-based features such as spectral, shape, and texture from images [15]. In contrast, this approach employs trained deep learning models to extract high-level features [108]. Following object-based feature extraction, a rule-based or machine learning-based classifier, such as Random Forest or Support Vector Machine, can be selected for classifying segments [6].

Compared to traditional features, deep learning-based features mitigate information loss resulting from manual operations and provide supplementary semantic information. Present research commonly utilizes semantic segmentation models as feature extractors [58], [169], comprising an encoder and a decoder, to output a feature map (like a heatmap) of the same size as the original input [108], [170]. For example, Ghorbanzadeh et al. [127] trained a model to detect landslides, where the output heatmap represents the probability of each pixel belonging to a landslide area. Additionally, Robson et al. [108] employed a CNN and Gaussian filter to generate a smoothly varying glacier probability heatmap. Subsequently, they utilized the probability heatmap as an additional feature for classification. Therefore, this approach may be suitable for single-type recognition tasks, as it fully leverages deep learning's feature extraction capabilities for a specific category and the boundary description advantages of OBIA.

Deep learning-derived data can also serve as input for traditional unsupervised segmentation methods [37], [128], as the output of deep learning models can effectively represent regions of interest. Compared to segmentation solely based on images, this approach with deep learning-derived features may yield better results in complex environments or when extracting semantic objects such as city communities, as deep learning-derived data tends to contain less noise than the original images [37]. However, since semantic segmentation models often struggle with object boundaries (due to downsampling and upsampling structures), employing this approach might necessitate additional fine-tuning or the implementation of post-processing techniques to address issues with blurry boundaries [128]. For example, Ghorbanzadeh et al. [37] initially performed segmentation based on high-resolution images using Multi-Resolution Segmentation, followed by merging and classifying over-segmented objects using CNN-derived probability. Timilsina et al. [128] conducted post-processing, including the assign-merge function, pixel-based object resizing, and object removal, to refine tree objects after utilizing deep learning-derived output for Multi-Resolution Segmentation.

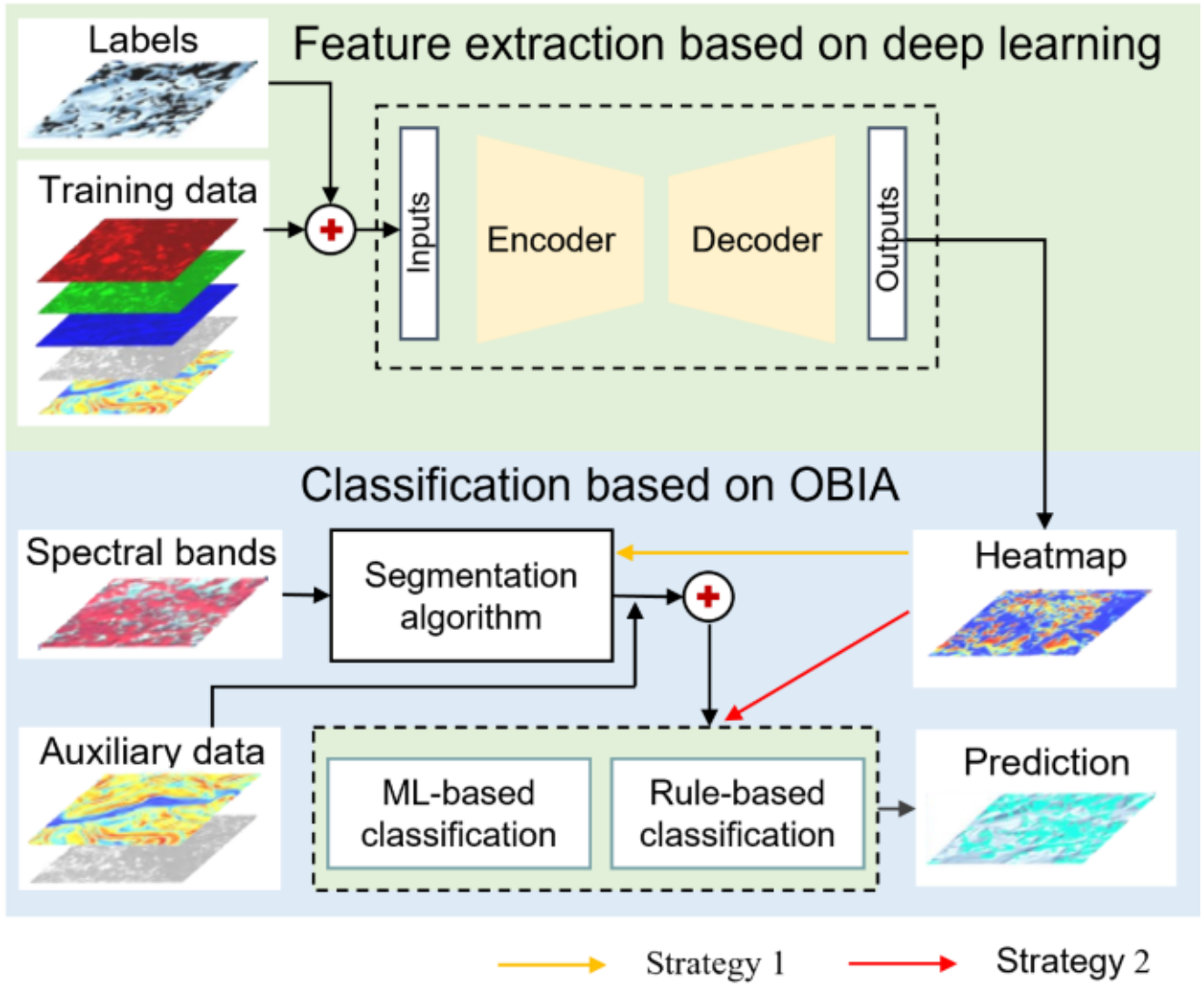

Fig. 4. Object-based image classification with deep learning-derived data [127].

*Identifying the land cover directly by semantic segmentation*

Semantic segmentation through deep learning represents a transformative approach to object-based image analysis. This innovative paradigm automates and consolidates the traditional OBIA processes of object segmentation, feature extraction, and object classification into a unified and efficient methodology for land cover classification (Fig. 5). It produces homogeneous image regions that correspond to image objects in traditional OBIA, each with designated classes.

Traditionally, OBIA groups pixels based on spectral and textural similarities using techniques such as FNEA or superpixel segmentation [5], [32], [171]. In contrast, semantic segmentation with deep learning automates and refines this process. Within a convolutional neural network (CNN), an encoder processes the input image through multiple convolutional layers, each layer applying filters that capture diverse features at varying scales [172], [173]. As the image advances through the encoder, spatial resolution decreases, causing each pixel in the deeper layers to represent a larger area of the original image. These pixels can be considered "objects," composed of clusters of original pixels, with segmentation implicitly achieved through the network's architecture rather than explicit rule-based region merging [174].

Traditional feature extraction, following object segmentation, computes descriptors at the object level from initial pixel-level features. In semantic segmentation, this step is integrated with segmentation; as the network segments the image into meaningful objects, it simultaneously extracts features. Multiple convolution kernels in the encoder are designed to detect different aspects of the input image (e.g., edges, textures, colors), extracting features at multiple scales and aggregating them in the deeper layers of the network [172], [175]. This provides a rich and contextually aware feature set that comprehensively describes each segmented object.

In traditional OBIA, after segmentation and feature extraction, classification algorithms categorize each object into predefined classes. Semantic segmentation integrates this classification step directly into the workflow. The decoder part of the network upsamples the feature-rich and lower-resolution output of the encoder back to the original image size, refining object boundaries and applying classification labels to each pixel [172], [175]. This process is typically supervised, using labeled training data to teach the network the correct class for each object based on its features.

Following the aforementioned innovative paradigm, researchers have applied a variety of deep learning models for semantic segmentation, such as Unet [176], SegNet [177], ResUNet-a [174], SSAtNet [178], SegFormer [179], and UNetFormer [180], and CLCFormer [181], to land cover classification from remote sensing images instead of the conventional OBIA. Review papers by Cheng et al. [182] provide more information on semantic segmentation using deep learning models for land cover classification. Moreover, as noted in Section III C, some researchers have incorporated strategies from OBIA into the design of semantic segmentation models to address issues like boundary blur in semantic segmentation results. For instance, incorporating shape and boundary information, as used in OBIA-based urban land cover mapping Li et al. [183], into multi-task semantic segmentation models improves farm field extraction Long et al. [184]. This also suggests that OBIA concepts hold potential for further developing semantic segmentation models.

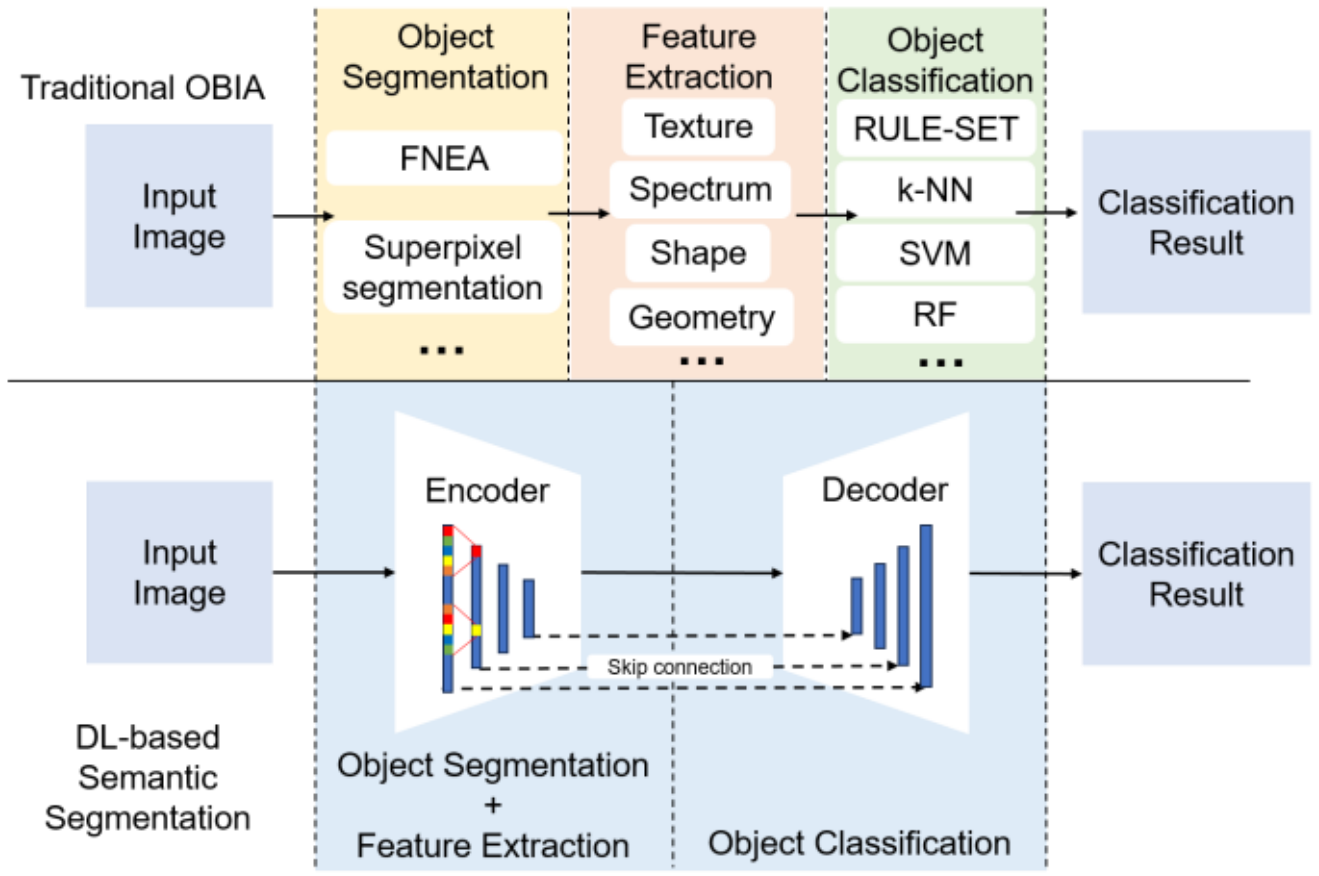

Fig. 5. Traditional OBIA-based land cover classification approaches and DL-based semantic segmentation approaches.

### C. *Pixel-based classification by deep learning and then refine the classification by segments*

This section outlines a hybrid strategy that combines pixel- or patch-level deep learning classification and object-level post processing (Fig. 6), culminating in the integration of results from both approaches. However, for the classification step, most of these studies employed pixel-based classification, except for Tong et al. [185] who used a patch-based method. This strategy capitalizes on the precision of Object-Based Image Analysis to delineate boundaries and the high accuracy label of the pixel-based method due to deep learning, while minimizing the speckled noise often associated with pixel-level processing.

The core methodology involves conducting the classification workflow predominantly at the pixel level, with object-level processing being confined to image segmentation. Classification within objects is then predominantly determined through a majority voting mechanism that assigns the label based on the most common outcome among the pixels. Early implementations of this strategy often used traditional classification algorithms. For instance, Costa et al. [186] applied a Decision Tree classifier at the pixel level and integrated this with object segmentation from the Multi-Resolution Segmentation (MRS) algorithm, using majority voting for final label assignment. Similarly, Li et al. [187] utilized a pixel-level Maximum Likelihood Classifier (MLC) with an advanced watershed algorithm for segmenting multispectral images, assigning labels based on the predominance of pixel classifications within each segment.

In more recent applications, semantic segmentation technologies have become the preferred approach for pixel-level classification. For example, Mboga et al. [188] employed a Fully Convolutional Network for pixel classification, paired with region-growing segmentation for object delineation, and used majority voting for object classification. Sun et al. [189] developed a custom multi-filter CNN to address the limitations of single-scale CNNs' receptive fields, integrating it with OBIA processes utilizing the MRS segmentation technique. Additionally, Liu et al. [13] developed a hybrid method that combines Object-Based Post-Classification Refinement (OBPR) with CNNs. This strategy ensures accurate pixel-level classification while object-level processing provides robust object units, culminating in integration via voting. Notably, despite semantic segmentation models' challenges in capturing fine boundaries accurately, this integrated approach shows promise. Pan et al. [190] further demonstrated that combining semantic segmentation with OBIA not only addresses the heterogeneity issues present in traditional Object-Based CNN (OCNN) classifications but also surpasses earlier process frameworks in performance.

Furthermore, integration methods extend beyond simple majority voting. Du et al. [191] used Conditional Random Fields (CRF) to integrate pixel-level classification probabilities with object units, incorporating additional road network and deep learning features to enhance method extensibility. Additionally, Zhao et al. [130] combined deep learning-generated features with pixel-level object features to retrain a two-layer Multi-Layer Perceptron (MLP) for object label determination. Zhang et al. [192] introduced a Joint Deep Learning (JDL) model that integrates an MLP with a CNN, employing a Markov process for iterative updates in Land Use (LU) and Land Cover (LC) classification. These innovative approaches suggest potential for significant enhancements over traditional voting methods, although more empirical evidence is needed to confirm these benefits.

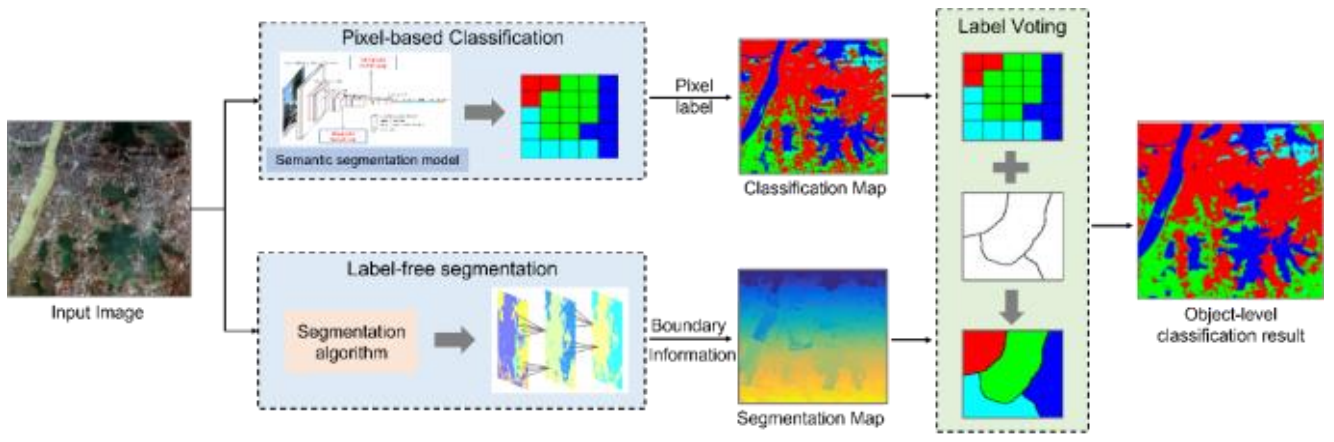

Fig. 6. The integration strategy by pixel- or patch-level deep learning classification and object-level post processing [185].

### D. *Graph-based deep learning for OBIA*

Traditional OBIA analyzes objects individually without considering the context provided by neighboring objects. To address this limitation, graph-based approaches have been introduced for OBIA. A graph consists of nodes and edges, where each node represents an object, and edges connect neighboring objects [148], [193] (Fig. 7). In object-based time-series analysis, edges can also connect objects from consecutive time steps if they overlap spatially [194].

Once the graph is constructed, various algorithms can analyze it for regression or classification tasks at the node, edge, or graph level. Land cover mapping can be treated as a node classification task using graph analysis methods. Graph analysis can serve as postprocessing for standard OBIA results to improve map quality. For example, Liu et al. [193] designed a fully learnable conditional random field (CRF) to enhance object-based land cover mapping accuracy. This method constructs a graph for each object and controls the influence of neighboring objects using a compatibility matrix and weighted feature similarity, both of which are learnable from training data. This CRF improved classification accuracy by 6% to 16% depending on the classifier and model inference choice.

Additionally, several algorithms can directly train the graph in an end-to-end manner to classify nodes, including Graph Convolutional Network (GCN) [195], GraphSAGE (GSAGE) [196], Graph Attention Network (GAT) [197], ResGatedGCN (ResGGCN) [198], and Graph Transformer (GT) [199]. Most of these algorithms are implemented in the PyTorch Geometric package [200]. Benchmark studies show that graph-based algorithms outperform random forests, with GT showing the best performance for object-based land cover mapping using remote sensing time series data [194]. The superiority of graph-based approaches is also evident in other studies compared to traditional OBIA using random forests or SVM [92], [148]. Moreover, combining the context information provided by graphs with features extracted from convolutional neural networks (CNN) can achieve higher accuracy than deep learning-based semantic segmentation models like UNet [201].

Graph-based methods require input features for each node (i.e., object). Features used in traditional OBIA, such as spectral, texture, and geometry features, can also be used for graph analysis [193]. An added advantage of graph analysis is that feature extractors like CNNs can be integrated with the graph to automatically learn and extract the best features from objects in an end-to-end training process [92], [202].

While graphs used for OBIA are usually undirected and analyzed at the node level, directed graphs are useful for certain applications, especially in time-series analysis where graph-level analysis is required. For example, Khiali et al. [147] used directed graphs for satellite image time series (SITS) data to depict dynamic changes in objects over time and applied clustering algorithms on graph-level features to identify objects with similar temporal behavior. Building on this, Debusscher et al. [203] proposed an alternative method for flood detection using SITS data.

Graph analysis could become a powerful tool for Object-Based Image Analysis (OBIA) in the future due to its enhanced accuracy, advanced time-series analysis capabilities, and integrated feature learning. Future studies should explore additional functionalities of graph analysis for OBIA beyond node classification, such as edge feature embedding and classification, and node-level regression. Additionally, computational efficiency should be compared between graph-based OBIA and semantic segmentation models to evaluate whether OBIA could provide computational benefits in addition to its higher accuracy.

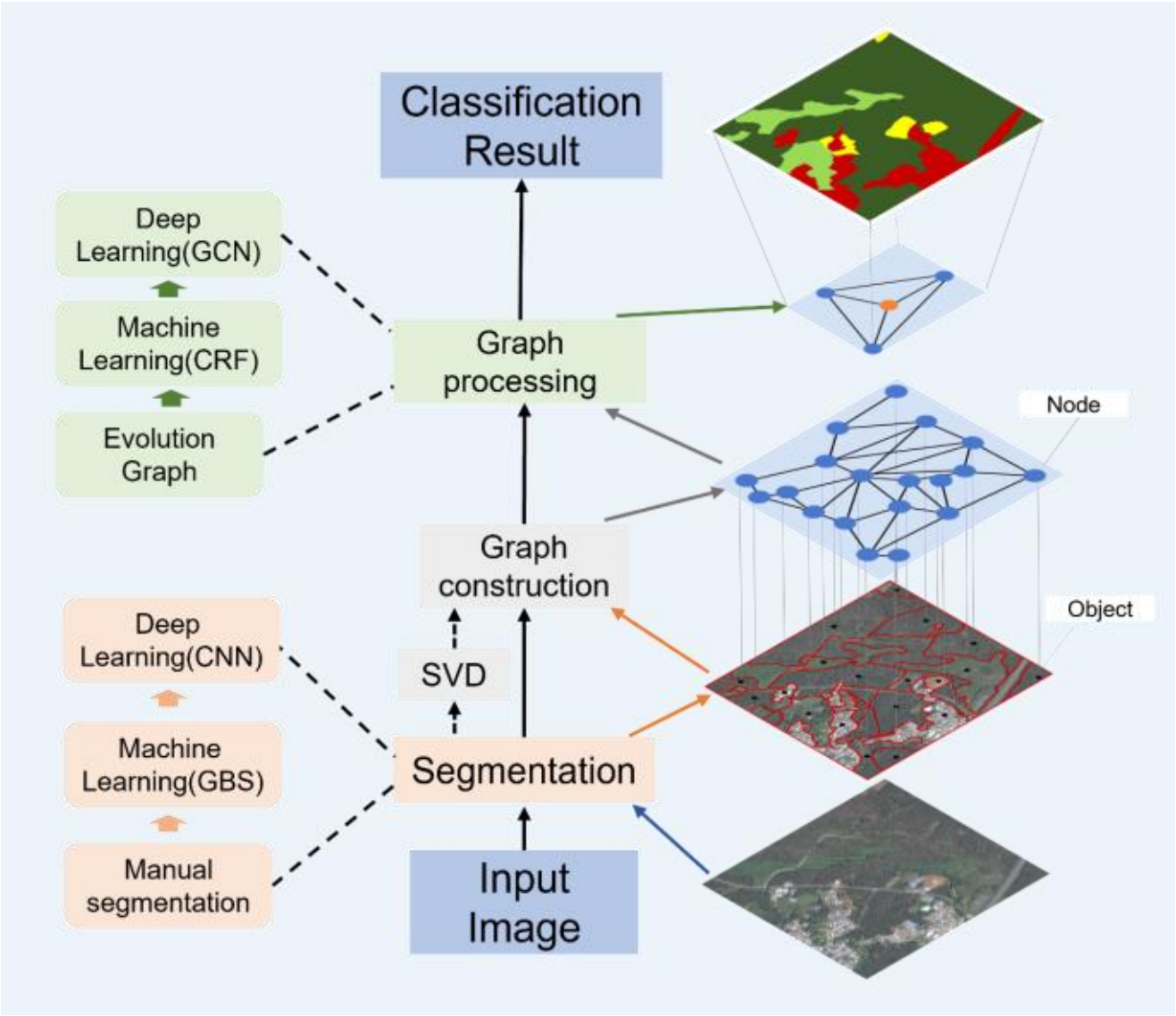

Fig. 7. A general graph-based OBIA framework.

## V. Deep learning for object-based time-series analysis

Compared to traditional object-based change detection methods that rely on handcrafted features, deep learning models can extract complex and high-dimensional features from remote sensing images, facilitating fine-grained detection of changes in ground objects [204], [205]. Post-classification change detection is a classical approach that shifts the task of change detection to a classification problem. This method employs deep learning models to classify images from different time periods and then compares the classification results to identify changes. For instance, Abbasi et al. [206] utilized very high resolution (VHR) imagery to create a series of image patches for roof objects at seven different time points. They combined DenseNet121 and LSTM models to classify each image patch in the sequence, generating a change map by analyzing the transitions in labels over time. A significant advantage of this approach is its independence from the need for radiation normalization, allowing it to handle data acquired under varied collection conditions. However, it depends heavily on the accuracy of image classification; any misclassification can be magnified in the final change detection map.

An alternative approach involves extracting feature vectors from images of different time periods, emphasizing change features through feature fusion, and ultimately generating a change map based on deep change features [207]. For instance, Wang et al. [208] proposed a method using a Siamese network that processes images from two periods through twin sub-networks with shared weights, extracting homogeneous features [209]. After fusing the feature vectors, they are inputted into a change decision network to detect pixel changes, and then FNEA is used to refine change detection results at the object level by implementing pixel voting within the same object. Although this method produces a binary change map (changed vs. unchanged), it lacks semantic information about the changes [210]. Liu et al. [142] improved on this by feeding the fused feature vectors into a classifier to obtain multi-class change-type results. As illustrated in Fig. 8, their method first employs the simple linear iterative clustering (SLIC) algorithm for image segmentation and then converts the segmented objects into patches. These patches, from two time periods, are then processed through two independent CNNs to extract feature vectors, which are then fused using methods like concatenation, differencing, and LSTM. The integrated features are subsequently inputted into a classifier to generate "from-to" change labels. This approach is beneficial for generating semantic change maps, although constructing the "from-to" training dataset can be challenging, particularly for large-scale change detection tasks.

The aforementioned methods typically follow a two-step process: first, employing traditional segmentation methods to accomplish object extraction tasks, and subsequently, utilizing deep learning models for change detection tasks [73]. In recent developments, some studies have adopted a multi-task learning strategy that integrates segmentation and change detection into a single deep learning framework. This approach enables the interaction between object features and change features, thereby improving the accuracy of change detection. For example, Gao et al. [211] proposed an end-to-end building change detection framework that uses both sub-networks for extracting building masks and rough change masks, respectively. The rough change masks generated are then refined at the object level based on precise building extraction results. Similarly, Zheng et al. [143] combined object localization networks with change detection networks to assess the extent of building damage post-disaster. This methodology allows deep object features to steer the classification of damage, yielding a map of building objects alongside a pixel-level map that indicates the degree of damage. Subsequently, to maintain semantic consistency across each building object, object-based post-processing is applied.

Graph-based strategies have become more popular in object-based change detection in recent years because they are good at getting and using object-specific spatial context information, as shown in studies by Wu et al. [212] and Guttler et al. [154]. Particularly, graph convolutional networks are adept at modeling interactions among unstructured data within deep learning frameworks. Long et al. [213], for example, used a semantic segmentation network to get bi-temporal semantic features of ground objects and then GCNs to look at how these semantic features and their differential attributes related to each other. This method effectively capitalizes on the significance of each feature layer and its interaction data to improve change detection accuracy. Future research could look into integrating segmentation models with graph-based approaches to advance deep learning in object-based change detection.

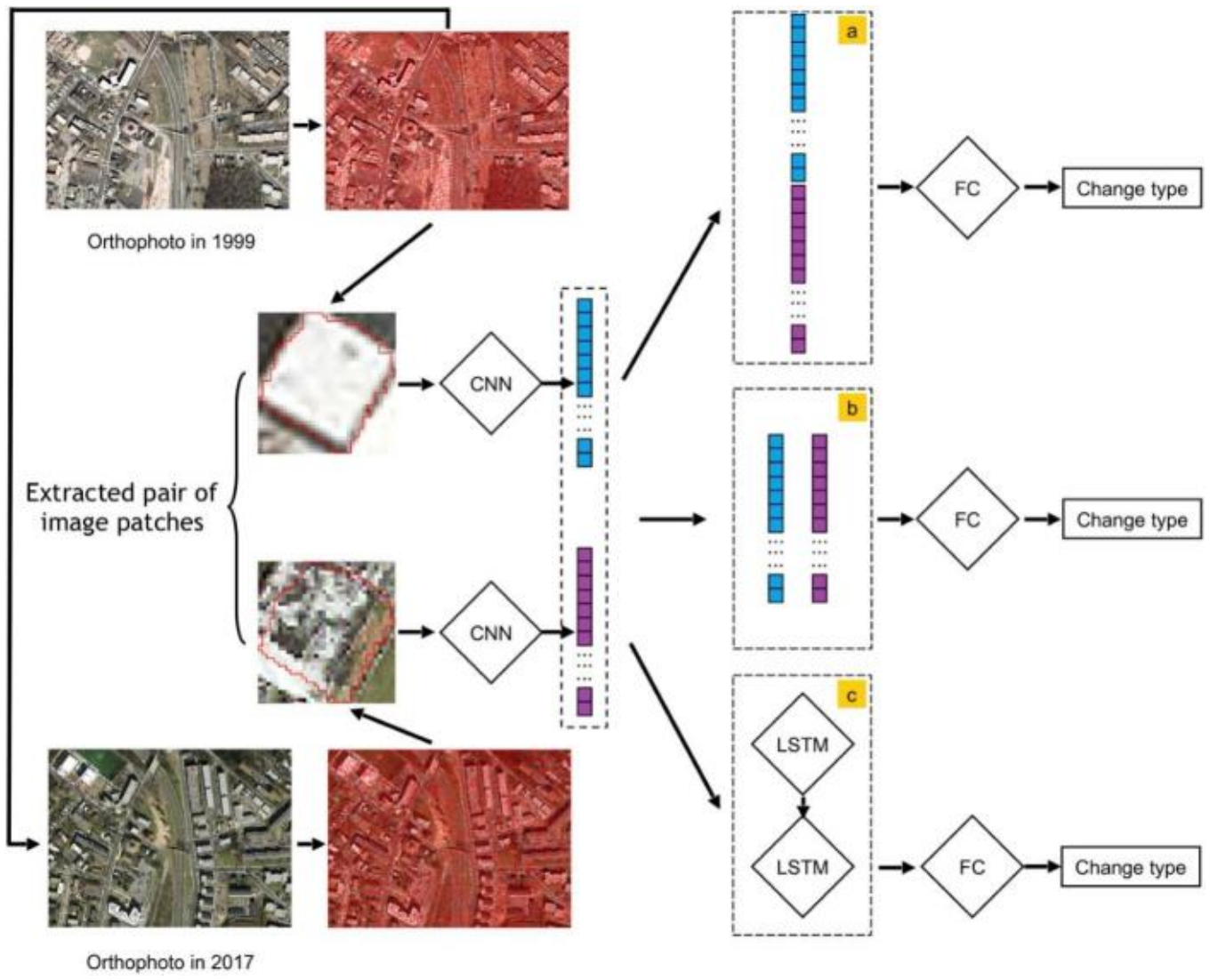

Fig. 8. The architecture of deep learning models for object-based change detection proposed in Liu et al. [142].

## VI. Deep Learning for Object-Based Time-Series Analysis

The surge in deep learning technologies has significantly heightened interest in their applications for remote sensing SITS (Satellite Image Time Series) analysis, as noted by Miller et al. [214]. In the realm of object-based SITS analysis, there has been a progression from the initial use of Convolutional Neural Networks [215] to Recurrent Neural Networks [216] and, more recently, to Graph Neural Networks due to its ability to capture contextual information of objects by constructing regional adjacency graphs [92]. Therefore, for Satellite Image Time Series analysis with deep learning, a main challenge is to design methods able to leverage the complementarity between the temporal dynamics and the spatial patterns [92]. This requires developing models that not only capture the intricate temporal sequences but also the spatial and contextual information of the segmented objects.

CNNs have been particularly successful across various domains due to their superiority in capturing local spatial patterns, making it the first deep learning model to be considered for processing object-level time series data. For example, Li et al. [167] proposed a novel Temporal Sequence Object-based Convolutional Neural Network (TS-OCNN), classifying the agricultural crop type using a standard CNN classifier from image time-series at the object level, where segmented objects were obtained through the Multi-Resolution Segmentation (MRS) algorithm. However, CNNs may struggle to capture global contextual information and long-term series information, leading researchers to develop CNN - based models. For example, Censi et al. [92] combined CNNs with the SLIC algorithm for segmenting images from time series, integrating contextual information between objects through a graph attention mechanism to account for both temporal dynamics and spatial context, thereby achieving high-precision land cover mapping. Furthermore, Ienco et al. [215] utilized multiple CNN blocks to extract detailed information from different object components, enhancing classification accuracy through an attention mechanism that addresses the challenge of limited label data.

However, this model, RNN, addresses CNN's limitations in handling sequence data by effectively capturing temporal dynamics and long-term dependencies within time series data. For example, Ienco et al. [217] utilized a dual RNN model (2RNN) to classify segmented objects with 24 Sentinel-1 (S1) images and 34 Sentinel-2 (S2) images, thereby enhancing spatial and temporal dependencies and improving land use/cover and change detection efficiency. Following Ienco et al. [217], Gbodjo et al. [216] developed an object-based SITS mapping architecture based on an extended RNN model adapted for the specificity of SITS data through an enhanced attention mechanism (Fig. 9). Despite these successes, RNNs for SITS analysis still face challenges, such as gradient issues and difficulties in processing long time series. This led to the development of improved variants like Long Short-Term Memory (LSTM) networks, which incorporate a control mechanism to better handle long sequences and their dependencies [145]. Consequently, LSTMs continue to be explored for their robustness in long-time series classification and mapping, although challenges such as computational complexity and interpretability persist [144], [145]. On the other hand, Graph Neural Networks (GNNs) offer a solution by constructing regional adjacency graphs for segmented objects, thereby maximizing the use of contextual information between segments. For example, Khiali et al. [147] introduced the EGraphClustering framework, utilizing a graph-theoretical representation to capture the evolutionary attributes of spatiotemporal entities for clustering tasks, underscoring the potential of GNNs in SITS analysis.

The Transformer architecture, known for its superior performance in adaptive and global information aggregation, has shown significant promise for SITS applications [74]. Chen et al. [218] integrated OBIA with the visual Transformer architecture, significantly reducing the computational load of self-attention modules. Furthermore, some studies [74], [204] in the computer field have attempted to utilize an integrated framework of GNNs and Transformers for SITS analysis, while their applications within the OBIA domain remain blank. Recently, the emergence of the Mamba architecture, which mitigates the global information capture limitations of CNNs and the computational intensity of Transformers, has shown great potential in dense prediction tasks of remote sensing imagery [219], [220]. It is suggested that future research should further explore the potential of GNNs, Transformers, and Mamba models to advance the field of object-based SITS analysis.

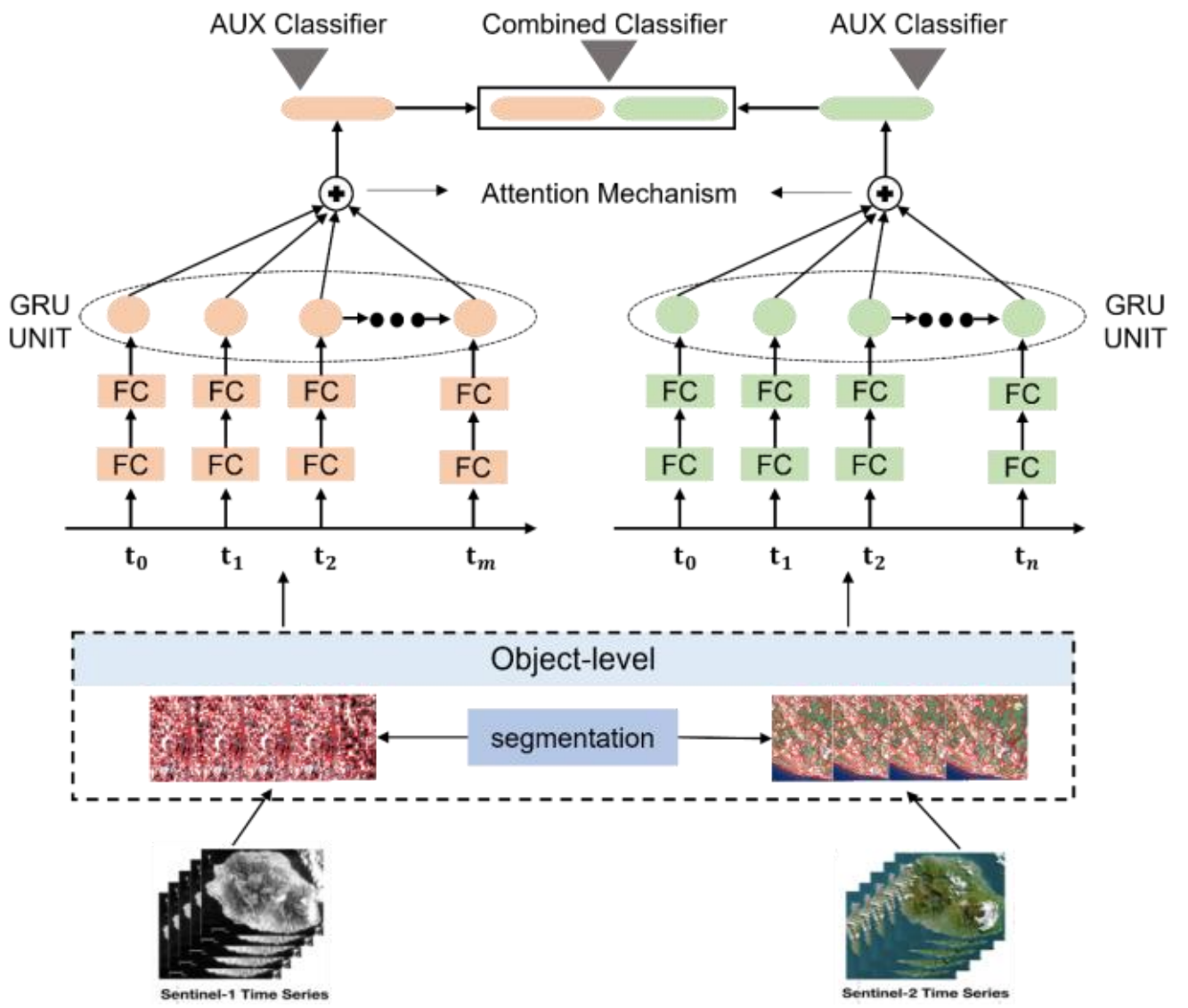

Fig. 9. OD2RNN Deep Learning Architecture [216], [217].

## VII. SEGMENTATION WITH DEEP LEARNING FOR OBIA

Traditional segmentation algorithms, such as edge-based and region-based segmentation, along with hybrid methods that combine both, have shown promise but face challenges like optimal parameter selection, segmentation optimization, and handcrafted feature extraction [16], [221]. Conversely, deep neural networks have revolutionized image segmentation by automating network parameter learning and feature extraction driven by specific tasks [14]. To date, numerous deep neural networks have been designed for remote sensing image analysis, especially for semantic segmentation. The Fully Convolutional Network (FCN), introduced by Long et al. [155], is a notable model that substitutes fully connected layers with convolutional ones. This adaptation has led to its wide application in tasks like road segmentation [222], building extraction [223], [224], and land cover classification [225]. Significant developments following FCN include U-Net [176], distinguished by its symmetric U-shaped architecture; SegNet [226], employing a unique storage pooling index mechanism; and DeepLab [227], introducing the concept of atrous convolution. These models facilitate simultaneous pixel-level segmentation and classification [228]. Despite their achievements, these semantic segmentation models often struggle to precisely capture the boundary details of ground objects and require a substantial number of training samples. This limitation impacts their ability to accurately delineate the outlines and spatial relationships of ground features, which can hinder the interpretation of complex scenes [229].

To overcome these limitations, some studies have integrated traditional OBIA frameworks with semantic segmentation models to enhance semantic classification. For instance, Du et al. [191] utilized a semantic segmentation network for pixel classification, followed by object segmentation using Multi-Resolution Segmentation (MRS) and category identification through pixel voting. The classification results were further refined using a road-constrained Conditional Random Field (CRF) that considers spatial context information (Fig. 10a). Some studies also directly improved the semantic segmentation process to enhance the boundary; for example, Li et al. [158] integrated an edge detection branch stream into the semantic segmentation branch stream to compensate for boundary loss in semantic segmentation.

In addition, the Segment Anything Model (SAM) has recently attracted significant attention in remote sensing for its excellent generalization and zero-shot segmentation capabilities [230], [231]. Osco et al. [232] demonstrated SAM's effectiveness with minimal training samples, and Hu et al. [233] further assessed SAM's performance in an Object-Based Classification (OBC) task. For this study, images were first segmented by SAM, followed by the extraction of geometric, textural, and spectral features from the segmented objects. These features were then classified using a random forest, yielding classification results that surpassed traditional OBC methods (Fig. 10b) [233]. This highlights SAM's potential as a promising alternative to traditional segmentation algorithms within the OBIA paradigm, achieving accurate segmentation results with minimal manual input or intervention.

Additionally, instance segmentation technique, as an important branch task in the field of image segmentation, has been actively explored in remote sensing. Unlike semantic segmentation, which fails to differentiate objects within the same category, instance segmentation [234] uniquely identifies each instance of ground objects, similar to tasks in object detection in computer vision. In remote sensing image analysis, researchers have successfully applied instance segmentation techniques to distinguish individual buildings [235] and identify different objects within the same type [236], [237], particularly in the mission of counting trees [38], [39]. Li et al. [39] established a multitask deep learning-based network to jointly solve the individual tree counting and crown segmentation tasks, as shown in Fig. 10c. Sun et al. [236] utilized Cascade Mask R-CNN, a popular instance segmentation method, to count trees in Guangzhou. They segmented the input images into patches and used a Region Proposal Network (RPN) to generate proposal regions (anchor boxes) for individual trees. A three-stage cascaded detector then determined the presence of trees in these regions and refined their boundary delineations. Also, Wu et al. [237] count large herds of migratory ungulates (wildebeest and zebra) using a U-Net-based ensemble model. While panoptic segmentation [36], which combines the ideas of semantic and instance segmentation to label pixels in great detail and identify objects, is still not fully explored in remote sensing [238].

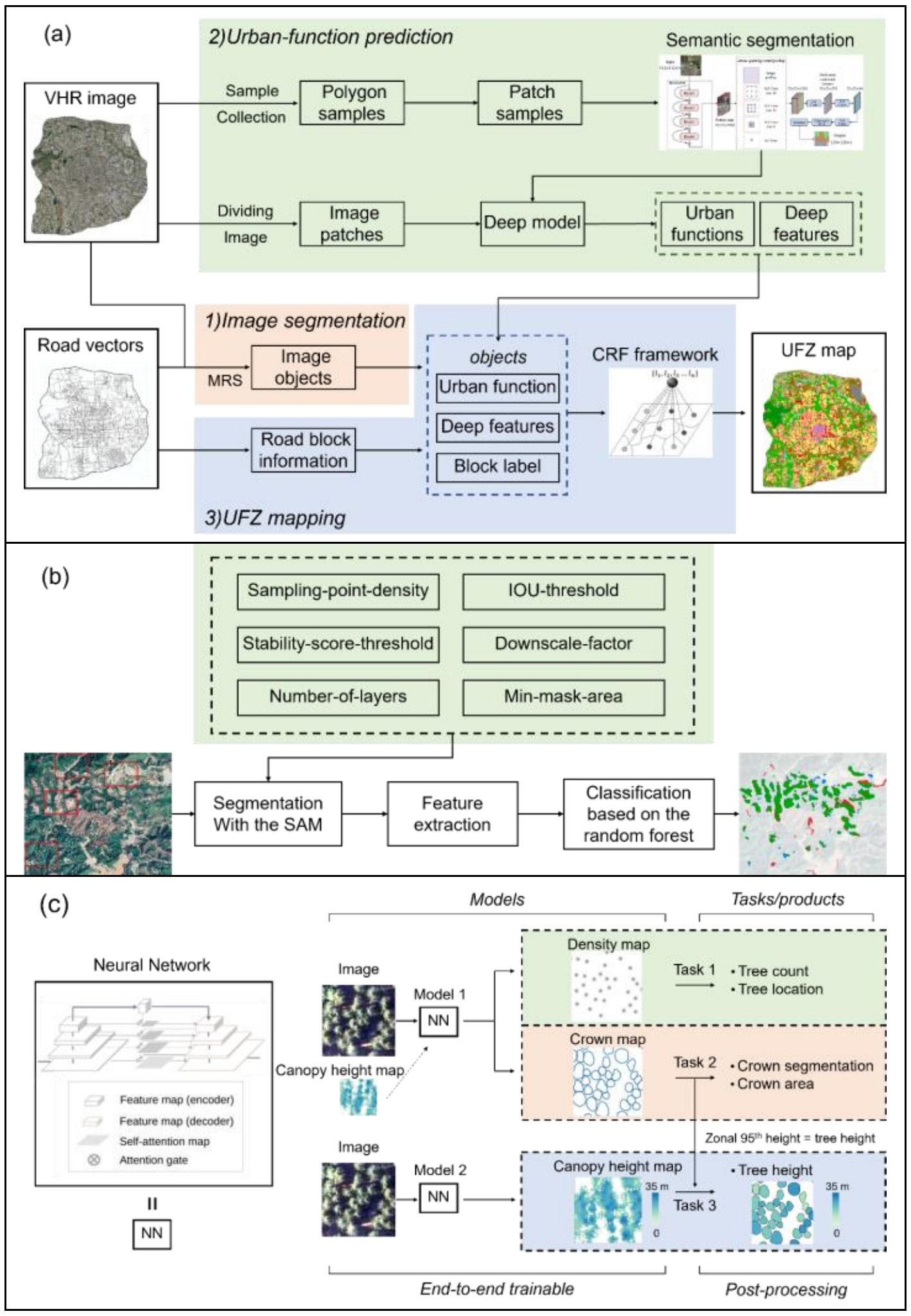

Fig. 10. Three applications of relevant deep learning image segmentation models. (a) semantic segmentation with CRF [191]. (b) SAM [233]. (c) instance segmentation [39].

## VIII. The Fields of Applications Combining Deep Learning and OBIA

This section provides a well-rounded review of the application fields of various DL methods with OBIA from the searched data in Fig. 1, including land cover and land use (LULC) classification, natural environment detection, and urban environment detection (Fig. 11). Next, we analyze and summarize the current application and development trends for each field.

Land cover/land use (LULC) classification as the fundamental mission of OBIA is pivotal as it depicts ecological environments and human activities on the Earth's surface, serving as essential support for remote sensing studies and land management [239]. The OBIA paradigm, enhanced with advanced DL algorithms, has become the mainstream of innovative solutions to achieving precise LULC classification. For example, Convolutional Neural Networks (CNNs), by leveraging their ability to capture intricate hierarchical structures, have significantly boosted classification performance [162], [240]. However, the scarcity of training samples often limits CNN effectiveness, leading to the adoption of transfer learning and active learning strategies to expand training datasets [241], [242]. Additionally, the integration of Recurrent Neural Networks (RNNs) with attention mechanisms is being explored to better handle multi-temporal remote sensing data, aiming to further enhance LULC classification accuracy [216].

Natural environment detection benefits more from OBIA approaches because of the greater spectral homogeneity of natural objects [136], [243]. Meanwhile, natural objects are typically larger and more regularly shaped, easily processed into fixed and consistent input data suitable for DL algorithms [47]. This compatibility extends the integration of OBIA and DL technologies to various natural environment detection tasks, such as ecosystem services detection [131], vegetation classification [128], [244], water detection [136], [243], crop mapping [245], coastline extraction [47], and desert detection [246].

The detection of urban environments has attracted increasing interest due to its direct linkage to human activities. Urban settings, however, present a challenge for OBIA due to their strong spectral heterogeneity and the fragmented, small-scale nature of urban objects [247], [248]. Deep learning commonly addresses these challenges by transforming spectral data into high-level features and spatial relationships, thereby facilitating the differentiation of urban objects from their surroundings. Convolutional Neural Networks, for instance, have proven effective in this regard [249]. As deep learning algorithms continue to evolve, their application in detecting fragmented urban environments is increasingly viable, making urban environment detection a significant area of OBIA applications. This includes building detection [249], [250], [251], urban landscapes [252], and road extraction [247]. Moreover, for urban landscapes, novel and sophisticated urban classification systems such as urban functional zones (UFZ) [191] and local climate zones (LCZ) [253] are gaining attention. These systems rely on deep learning to extract high-dimensional features for more accurate recognition, while minimizing reliance on the spectral properties of individual pixels.

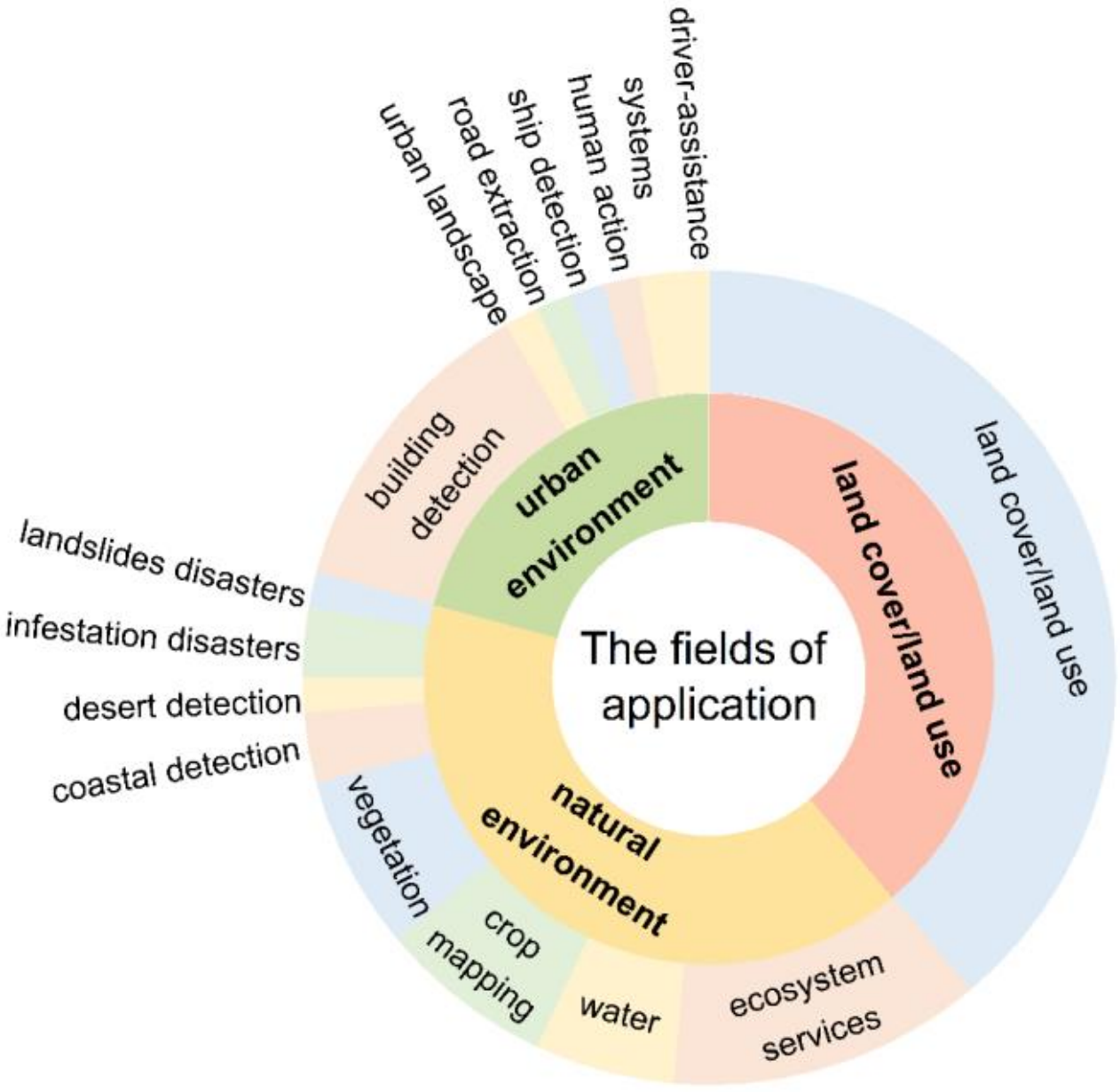

Fig. 11. The application fields using deep learning and OBIA.

## IX. Perspectives and Conclusions

This article reviews the state of the art in an important and underexploited research field: deep learning in OBIA. It summarizes the general tasks of OBIA and explores how deep learning can be implemented for these tasks. For the first time, several integration classification strategies for combining OBIA and DL are summarized for the OBIA research community. Additionally, the article reviews the application of DL in cutting-edge OBIA domains such as change detection, time series analysis, and segmentation.

Despite early successes, the full exploitation of deep learning in OBIA is primarily limited by several factors: 1) unclear concepts, since DL can be applied at various stages of OBIA (e.g., segmentation, feature extraction, classification); 2) the challenge of irregular segmentation objects, which makes it difficult for DL to directly handle these unstructured units; and 3) high training sample requirements and poor interpretability inherent in deep learning.

In conclusion, it is acknowledged that the meeting of OBIA and deep learning might seem unfortunate, as deep learning has the potential to overthrow the traditional OBIA paradigm due to semantic segmentation. This, in some ways, actually hampers the development of traditional OBIA. This could also account for the decline in the publication volume of OBIA and DL in recent years (Fig. 1), as they do not encompass a significant number of segmentation tasks related to classification. On the other hand, if OBIA is considered a goal-driven paradigm, we must admit that deep learning has brought new vitality to OBIA. Therefore, clarifying the concepts and expanding the traditional research domains of OBIA is necessary, as many new related domains, like semantic segmentation for land cover mapping, fundamentally adhere to the OBIA philosophy by aggregating and classifying similar pixels, despite not being considered object-based.

Therefore, we should differentiate segmentation based on specific object-level operational strategies, not just as a step in traditional OBIA. With segmentation taking on new task attributes (e.g., semantic segmentation, instance segmentation, and panoptic segmentation), it now accomplishes many remote sensing tasks without separate individual classification steps, such as semantic segmentation for classification and instance segmentation for counting interested objects (e.g., trees). Future research should therefore attribute more roles to segmentation in the field of OBIA, not only segmentation for segments.

Furthermore, traditional OBIA typically faces segmentation scale optimization and feature extraction issues, but the use of deep neural networks has revolutionized the image segmentation process, making these critical issues seem less significant due to the multi-level feature extraction capabilities of deep learning. However, semantic segmentation faces new issues of boundary blur, making traditional OBIA still look very important, as it can utilize the advantages of high spatial resolution remote sensing images to retain detailed boundaries. Therefore, the development of deep learning seems to be overturning traditional OBIA, but OBIA can still offer unique value.

The paper provides an in-depth study of DL for object-based image classification. It summarizes five common strategies that integrate DL into the OBIC process at different stages, including segmentation, feature extraction, and classification. However, the application of DL in other OBIA tasks remains relatively limited, especially in multi-temporal change detection and time series analysis, where further development of related theories and methods could pave the way for tackling the inconsistency of segmented objects across different times.

Creating DL models that work with OBIA, like the graph-based deep learning model, could solve many of the problems that arise when DL models are combined with OBIA, such as the fact that analysis units don't always match up in the spatial and temporal dimensions. Specifically, the use of Graph Neural Networks addresses the limits of lacking contextual information between segments, while Recurrent Neural Networks are able to fully capture the information of intricate temporal sequences. We also anticipate applying the advanced DL model more across various OBIA domains to tackle the challenges posed by traditional DL models.

In addition, compared to pixel units, because segmentation results in an exponential decrease in the basic units that can be processed, and deep learning naturally requires a large number of samples to achieve better classification effects, sample expansion becomes another important issue that needs to be resolved for better integration of OBIA with deep learning.

Overall, we should expand the concept of OBIA and not confine it to traditional OBIA paradigms. Particularly in the context of deep learning, OBIA should be revitalized by developing new methods or expanding a new philosophy. It is expected that this review will inspire renewed interest and innovative approaches in the OBIA community, thereby enhancing the role of OBIA in the evolving landscape of remote sensing and deep learning applications.